\documentclass{article}
\usepackage[latin9]{inputenc}
\usepackage{array}
\usepackage{rotating}
\usepackage{float}
\usepackage{multirow}
\usepackage{amsmath}
\usepackage{graphicx}
\usepackage{setspace}
\usepackage{microtype}
\usepackage[unicode=true,
 bookmarks=false,
 breaklinks=false,pdfborder={0 0 1},backref=page,colorlinks=false]
 {hyperref}

\makeatletter

\providecommand{\tabularnewline}{\\}
\floatstyle{ruled}
\newfloat{algorithm}{tbp}{loa}
\providecommand{\algorithmname}{Algorithm}
\floatname{algorithm}{\protect\algorithmname}

\usepackage[accepted]{icml2021}
\usepackage{amsfonts}
\usepackage{subfig}

\icmltitlerunning{Group Fisher Pruning for Practical Network Compression}

\@ifundefined{showcaptionsetup}{}{%
 \PassOptionsToPackage{caption=false}{subfig}}
\usepackage{subfig}
\makeatother

\begin{document}
\twocolumn[
\icmltitle{Group Fisher Pruning for Practical Network Compression}
\icmlsetsymbol{equal}{*}

\begin{icmlauthorlist}
\icmlauthor{Liyang Liu}{equal,thu}
\icmlauthor{Shilong Zhang}{equal,ai}
\icmlauthor{Zhanghui Kuang}{st}
\icmlauthor{Aojun Zhou}{st}
\icmlauthor{Jing-Hao Xue}{ucl}
\\\icmlauthor{Xinjiang Wang}{st}
\icmlauthor{Yimin Chen}{st}
\icmlauthor{Wenming Yang}{thu}
\icmlauthor{Qingmin Liao}{thu}
\icmlauthor{Wayne Zhang}{ai,st,qy}
\end{icmlauthorlist}

\icmlaffiliation{thu}{Shenzhen International Graduate School/Department of Electronic Engineering, Tsinghua University, Beijing, China}
\icmlaffiliation{st}{SenseTime Research, Hong Kong, China}
\icmlaffiliation{ucl}{Department of Statistical Science, University College London, London, United Kingdom}
\icmlaffiliation{qy}{Qing Yuan Research Institute, Shanghai, China}
\icmlaffiliation{ai}{Shanghai AI Laboratory, Shanghai, China}

\icmlcorrespondingauthor{Wenming Yang}{yang.wenming@sz.tsinghua.edu.cn}
\vskip 0.3in]



\printAffiliationsAndNotice{\icmlEqualContribution} 

\def\ie{\textit{i.e.}}
\def\eg{\textit{e.g.}}
\def\etal{\textit{et al.}}
\def\wrt{\textit{w.r.t.}}

\begin{abstract}
Network compression has been widely studied since it is able to reduce
the memory and computation cost during inference. However, previous
methods seldom deal with complicated structures like residual connections,
group/depth-wise convolution and feature pyramid network, where channels
of multiple layers are coupled and need to be pruned simultaneously.
In this paper, we present a general channel pruning approach that
can be applied to various complicated structures. Particularly, we
propose a layer grouping algorithm to find coupled channels automatically.
Then we derive a unified metric based on Fisher information to evaluate
the importance of a single channel and coupled channels. Moreover,
we find that inference speedup on GPUs is more correlated with the
reduction of memory\footnote[6]{We use "memory" to denote the number of elements in the output feature maps of all layers. The code will be available at \url{https://github.com/jshilong/FisherPruning}.}
rather than FLOPs, and thus we employ the memory reduction of each
channel to normalize the importance. Our method can be used to prune
any structures including those with coupled channels. We conduct extensive
experiments on various backbones, including the classic ResNet and
ResNeXt, mobile-friendly MobileNetV2, and the NAS-based RegNet, both
on image classification and object detection which is under-explored.
Experimental results validate that our method can effectively prune
sophisticated networks, boosting inference speed without sacrificing
accuracy. 
\begin{figure}
\begin{raggedright}
\includegraphics[scale=0.21]{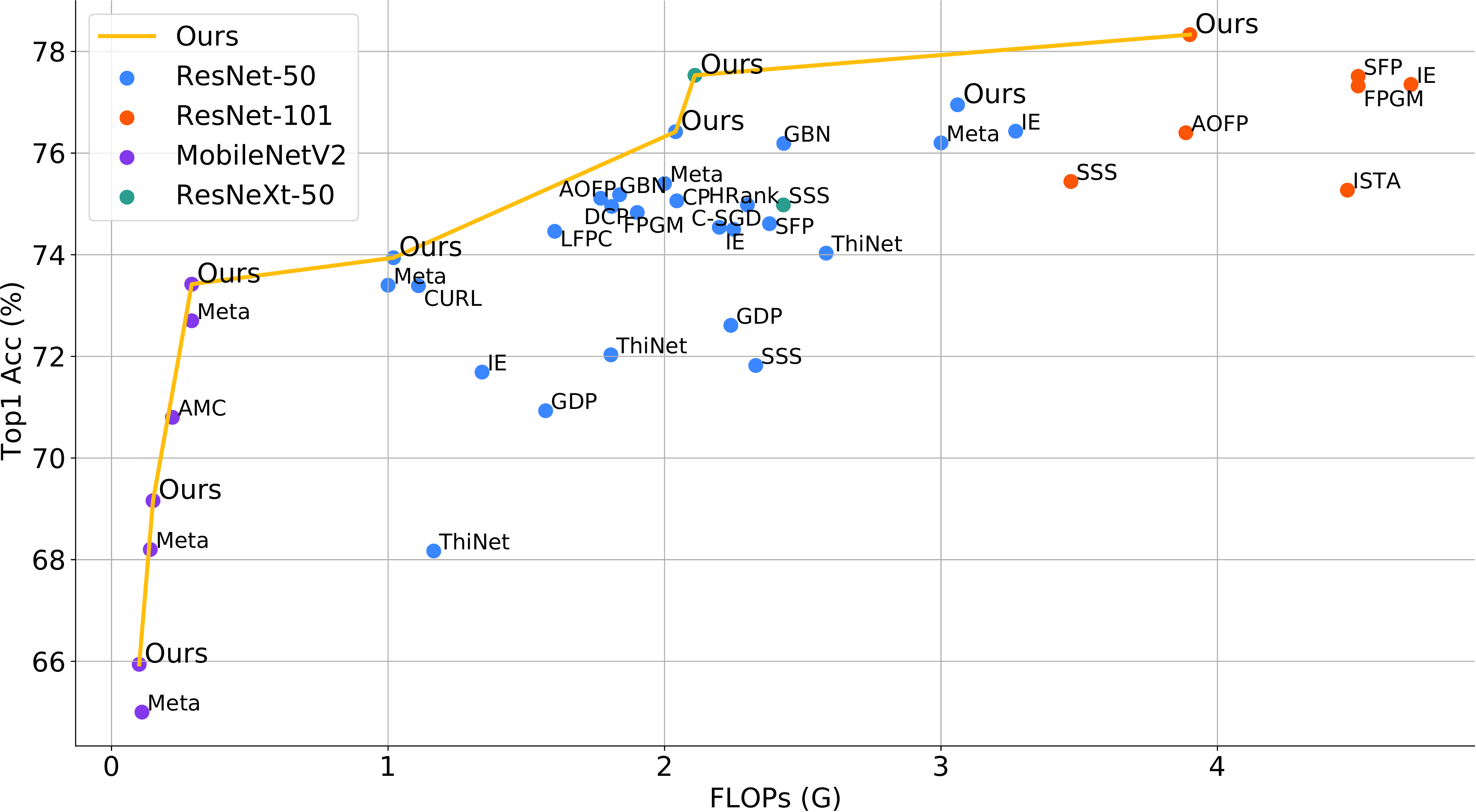}
\par\end{raggedright}
\caption{\label{fig:sota}{\small{}Compare top-1 accuracies of our pruned models
with state-of-the-arts under various FLOPs and network structures.
}}
\end{figure}
\end{abstract}

\section{Introduction}

Modern computer vision models equipped with deep networks exhibit
excellent performances in many tasks. However, they consume a great
amount of memory and computation during inference. It can hinder the
model deployment on edge devices where high-end hardwares are not
available. It can also limit the throughput of services on clouds,
resulting from considerable energy cost and inference latency. Network
pruning aims at increasing the inference efficiency with negligible
accuracy drop. It takes the trained dense model as input and prunes
weights or channels with little importances. Through fine-tuning the
pruned model can usually regain the lost performance caused by pruning.
\begin{figure}
\begin{raggedright}
\includegraphics[scale=0.21]{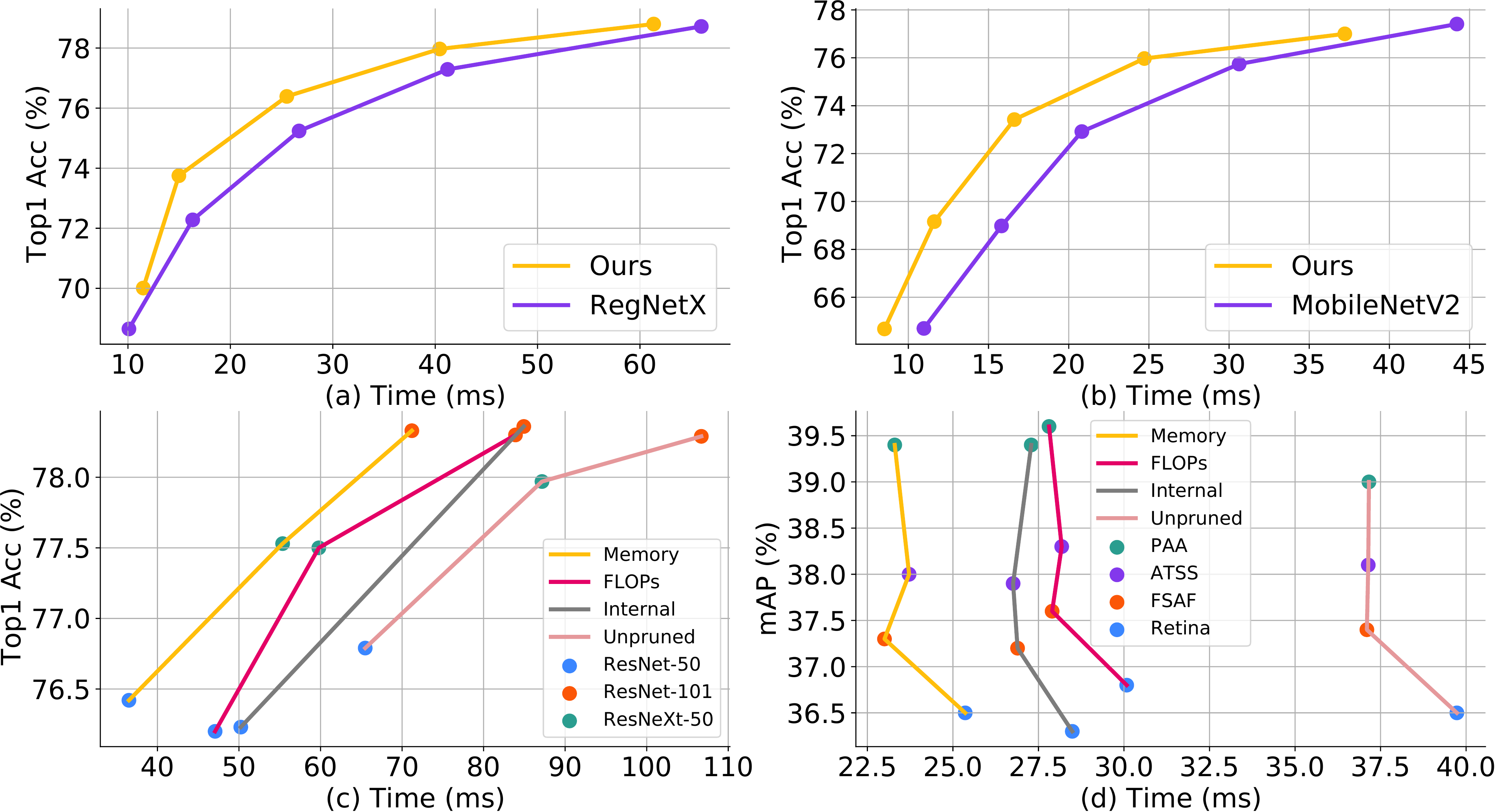}
\par\end{raggedright}
\caption{\label{fig:regnet}{\small{}Efficiency of our proposed method. In
(a) and (b) we compare our pruned models with the searched RegNet
and uniform-scaled MobileNetV2, respectively. In (c) and (d) }\emph{\small{}circles}{\small{}
with different colors represent different network structures, and
}\emph{\small{}lines}{\small{} show the results of pruning by different
strategies. ``Memory'' is our full method which normalizes importance
scores by memory reduction and prunes coupled channels via layer grouping.
``Internal'' only prunes the isolated channels. ``FLOPs'' normalizes
importances by FLOPs. ``Unpruned'' represents the unpruned model.
In (d) different detection networks are pruned and compared with the
baselines. The x-axis shows the inference time.
}}
\end{figure}

Although numerous \emph{channel pruning} methods have been proposed
in literature \cite{molchanov2016pruning,luo2017thinet,he2017channel,liu2017learning},
most of them study sequential networks such as AlexNet \cite{krizhevsky2017imagenet}
and VGGNet \cite{Simonyan15} where pruning the input channel of a
layer only affects the output channel of its single preceding layer.
However, recently developed networks are designed with complicated
structures such as residual connections in ResNet \cite{he2016deep},
group convolution (GConv) in ResNeXt \cite{xie2017aggregated} and
RegNet \cite{radosavovic2020designing}, depth-wise convolution (DWConv)
in MobileNet \cite{sandler2018mobilenetv2}, and feature pyramid networks
(FPN) \cite{lin2017feature} in object detection frameworks. These
structures have \emph{coupled channels} distributed in multiple layers,
which must be pruned or preserved simultaneously. Ignoring the coupled
channels and pruning them independently will definitely hurt the efficiency
in terms of both FLOPs (floating-point operations), memory access
and actual speedup during inference.

In this paper we propose a general framework named \emph{Group Fisher
Pruning }that can be applied to various complicated structures. Particularly,
we first introduce a binary mask initialized as $1$ for each input
channel. Then we propose a \emph{layer grouping} algorithm to automatically
find the coupled channels given computation graph of the network,
and we make the coupled channels \emph{share} the same mask. The importance
of a single channel is estimated by the loss change if it is discarded,
which is approximated by Fisher information and is proportional to
the squared mask gradient. Based on the single-channel importance,
we obtain the overall importance of coupled channels by the principled
chain rule of gradient computation. Pruning is done by iteratively
setting the mask of the least important channel to $0$, where the
coupled channels are pruned together. Finally the network is fine-tuned
to regain the lost accuracy. During fine-tuning and inference, the
channels with $0$ masks are explicitly excluded from the network,
and thus computation and memory cost can be practically reduced for
acceleration.

Moreover, we propose to normalize importances of channels by their
reductions of computation costs as we would like to prune the least
important channels with the most computation overheads to achieve
the best trade-off between accuracy and efficiency. However, we find
the commonly-used reduction of FLOPs is a rather biased estimator
for the actual inference speedup. In contrast, we propose to measure
the computational complexity of a channel by its reduction of memory
during pruning. Through experiments we find normalizing the channel
importance by the reduction of memory is more correlated with the
speedup than FLOPs in terms of the inference time on GPUs. 
\begin{figure*}
\begin{centering}
\includegraphics[scale=0.36]{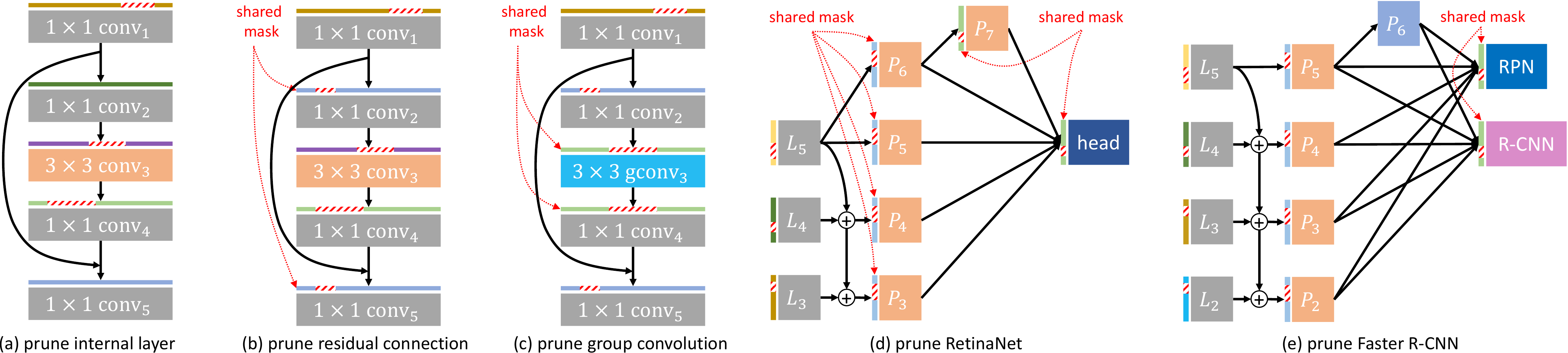}
\par\end{centering}
\caption{\label{fig:prune}{\small{}Prune different constrained structures.
For each Conv/FC layer we introduce channel-wise binary input masks,
where masks with the same color are shared and red stripes show the
pruned channels. Many previous methods only prune internal layers
in a residual block when residual connections exist as in (a), which
will lead to lower efficiency since only }\emph{\small{}output channels}{\small{}
of conv$_{2}$/conv$_{3}$ can be pruned but not those of conv$_{1}$/conv$_{4}$.
In contrast, we find coupled channels }\emph{\small{}automatically}{\small{}
by our proposed layer grouping algorithm and make them }\emph{\small{}share}{\small{}
masks as in (b), where conv$_{2}$ and conv$_{5}$ are found to have
coupled channels as they both are children of conv$_{1}$ in the computation
graph. We prune the coupled channels together such that the output
channels of conv$_{1}$/conv$_{4}$ can also be pruned correspondingly,
leading to higher efficiency. In (c) except for mask sharing in conv$_{2}$/conv$_{5}$,
the input and output channels of group convolution (gconv) are also
coupled and should be pruned simultaneously, which results in the
mask sharing of conv$_{3}$/conv$_{4}$. In (d) and (e) feature pyramid
and head networks for one-stage and two-stage detection frameworks
are shown, where the structures are more complicated than classification
networks. Note that P$_{6}$ in (e) is a pooling layer rather than
Conv layer, so there is no mask assigned to it. Even though, our method
is still able to achieve high efficiency via mask sharing for coupled
channels.}
}
\end{figure*}

Our proposed Group Fisher Pruning has the following advantages.\textbf{
Firstly}, it can prune any layers including those with coupled channels,
and thus achieves better trade-off between accuracy drop and actual
acceleration.\textbf{ Secondly}, it prunes globally rather than locally
\cite{he2017channel,luo2017thinet}, \ie, it obtains the pruning
ratio for each layer automatically without the cumbersome sensitivity
analysis of layer-wise pruning ratio \cite{yu2018nisp}, and thus
leads to higher accuracy. \textbf{Thirdly}, it estimates importances
of all channels in one pass via the principled Fisher information
instead of multiple forward passes for individual channels \cite{luo2020neural},
and thus is more efficient.  \textbf{Lastly}, in contrast with \cite{liu2017learning},
it does not depend on specific layers like batch normalization (BN)
and thus is more general so that we can prune more sophisticated structures
such as object detection networks, where such layers may not be na\"ively
adopted due to the larger input size. 

To demonstrate the generalization ability and effectiveness of the
proposed method to deal with complicated network structures, we conduct
extensive experiments on various backbones, including classic ResNet
\cite{he2016deep} and ResNeXt \cite{xie2017aggregated}, mobile-friendly
MobileNetV2 \cite{sandler2018mobilenetv2}, and recent NAS-based RegNet
\cite{radosavovic2020designing} on image classification (See Fig.
\ref{fig:sota}). We also evaluate our method on object detection,
which is more computation-intensive due to larger input image size
and more complicated network structure than image classification,
but rather under-explored (See Fig. \ref{fig:regnet}). 

Our main contributions
are: we introduce the concept of coupled channels, find them by the
proposed layer grouping algorithm, derive a unified metric to evaluate both coupled-channel and single-channel importances
(based on Fisher information), and normalize the importance by memory
reduction to realize higher speedup on GPUs without
sacrificing the accuracy.

\section{Related Work}

Network pruning can be generally categorized into unstructured and structured methods. Unstructured methods \cite{han2015learning,han2015deep,guo2016dynamic} prune unimportant weights in the model, but efficiency of the pruned sparse network can only be shown with the help of specialized libraries or hardwares. Recently, there are also efforts \cite{zhou2021learning,mishra2021acc} to develop N:M fine-grained sparse models, leveraging the innovations in general-purpose GPUs (\eg, NVIDIA Ampere architecture). On the contrary, structured methods prune the whole channels or filters with little importances, and thus actual speedup can be easily achieved without requiring sparse accelerators.

For structured pruning methods \cite{wen2016learning,lebedev2016fast},
different importance metrics have been proposed. PFEC \cite{li2016pruning}
employs $L_{1}$ norm of the channel weights, while SFP \cite{he2018soft}
uses $L_{2}$ norm of each filter. These methods rely on the ``smaller-norm-less-informative''
assumption \cite{ye2018rethinking} which may not be true especially
for structured pruning. CP \cite{he2017channel} and ThiNet \cite{luo2017thinet}
cast channel selection as reconstruction error minimization of feature
maps, where LASSO regression and greedy strategy are used to select
the pruned channels, respectively. However, they can only prune networks
in a layer-wise manner, as the least-square reconstruction happens
locally. NISP \cite{yu2018nisp} instead minimizes the reconstruction
error of the final response layer and propagates importance scores
through the entire network. The above methods also need sensitivity
analysis to decide the pruning ratio for each layer, which may be
time-consuming and sub-optimal. Network Slimming \cite{liu2017learning}
reuses BN layer scaling factors as importance scores so that channels
can be pruned globally. Although BN is prevalently used in image classification,
many applications such as object detection can not trivially adopt
BN because of the large input image size, which limits the application
scenarios of pruning methods based on BN scaling factors. SSS \cite{huang2018data}
introduces extra scaling factors to scale the outputs of various micro-structures
and solves the sparsity regularized optimization of scaling factors
by the accelerated proximal gradient method.

Apart from the heuristic-based importance evaluation methods, one
may use the exact loss change induced by removing a specific parameter
\cite{luo2020neural} to measure its importance, but it is prohibitively
expensive due to the large parameter number. Others try to approximate
the importance score via Taylor expansion on the loss. The seminal
work of OBD \cite{lecun1990optimal} and OBS \cite{hassibi1993second}
exploit the second-order derivative information to estimate the importances
of weights, but they may need to obtain the heavy-weight Hessian matrix
which is too large to compute for modern large-scale networks. L-OBS
\cite{dong2017learning} layer-wisely computes the Hessian matrix
to achieve tractable approximation. WoodFisher \cite{singh2020woodfisher} approximates
the inverse of Hessian matrix by the Woodbury matrix identity and
improves unstructured pruning based on OBD/OBS. PCNN \cite{molchanov2016pruning}
extends Taylor expansion to channel pruning and uses the first-order
information instead. It takes a greedy strategy to prune the least
important channels, interleaving pruning and fine-tuning. In place
of estimating importances of feature maps, IE \cite{molchanov2019importance}
applies Taylor expansion to the weights of a filter. These importance
estimation methods are more principled than magnitude-based ones,
but they seldom deal with structure constraints, for example, the
residual connections.

Besides the importances, another factor needed to be concerned is
computation, as we wish to prune the least important channels with
the most computation costs. Current methods \cite{molchanov2016pruning,theis2018faster}
typically add a regularization term to constrain FLOPs of the pruned
network. However, the same amount of FLOPs reduction may lead to different
actual speedups. Through experiments we empirically find that reduction
of memory access can act as a more accurate estimator for efficiency
gain, which is not explored in previous pruning methods. 

\section{Methodology}

We first introduce Fisher information \cite{theis2018faster} as single-channel
importance estimation, which can be used to prune channels in sequential
networks but can not deal with complicated structures. Then we propose
our layer grouping algorithm to find coupled channels in different
layers, and make the coupled channels share the mask so as to prune
them simultaneously. Finally we propose to use memory reduction as
importance normalization to achieve better trade-off between efficiency
and accuracy.

Given a training dataset $\mathcal{D}=\left\{ \boldsymbol{x}_{n},\boldsymbol{y}_{n}\right\} $
of image-label pairs and a network $\boldsymbol{W}_{0}$ trained on
it to convergence, we aim to prune the least important channel. We
introduce a binary mask (initialized as $1$) for each input channel
to achieve structured pruning, and one channel can be pruned by setting
its mask to $0$. During pruning, the input tensor $\boldsymbol{\mathsf{A}}\in\mathbb{R}^{n\times c\times h\times w}$
for a convolution (Conv) or fully-connected (FC) layer ($h=w=1$ in
FC) is element-wisely multiplied by the masks $\boldsymbol{m}\in\mathbb{R}^{c}$
with broadcasting to form the masked input $\boldsymbol{\widetilde{\mathsf{A}}}=\boldsymbol{\mathsf{A}}\odot\boldsymbol{m}$,
which is the actual input for each layer. During inference we explicitly
discard the channels with $0$ masks, both for a layer and its parents
in the computation graph, since pruning input channels of one layer
also prunes output channels of its preceding layers.

\subsection{Fisher Information Importance}

To evaluate the importance $s_{i}$ of a channel $i$, we apply Taylor
expansion to the network loss $\mathcal{L}$ and approximate the loss
change when discarding it (setting its mask to $0$):

\begin{align}
s_{i} & =\mathcal{L}\left(\boldsymbol{m}-\boldsymbol{e}_{i}\right)-\mathcal{L}\left(\boldsymbol{m}\right)\approx-\boldsymbol{e}_{i}^{\top}\nabla_{\boldsymbol{m}}\mathcal{L}+\frac{1}{2}\boldsymbol{e}_{i}^{\top}\left(\nabla_{\boldsymbol{m}}^{2}\mathcal{L}\right)\boldsymbol{e}_{i}\nonumber \\
 & =-\boldsymbol{e}_{i}^{\top}\boldsymbol{g}+\frac{1}{2}\boldsymbol{e}_{i}^{\top}\boldsymbol{H}\boldsymbol{e}_{i}=-g_{i}+\frac{1}{2}H_{ii},
\end{align}
$\boldsymbol{m}=\boldsymbol{1}$ is the all-one vector and $\boldsymbol{e}_{i}\in\mathbb{R}^{c}$
denotes the one-hot vector of which the $i$-th element equals $1$.
$\boldsymbol{g}\in\mathbb{R}^{c}$ is the gradient \wrt $\boldsymbol{m}$
and $\boldsymbol{H}\in\mathbb{R}^{c\times c}$ is the Hessian matrix.
As the model has converged and recall that $\boldsymbol{\widetilde{\mathsf{A}}}=\boldsymbol{\mathsf{A}}\odot\boldsymbol{m}$,
$\boldsymbol{m}=\boldsymbol{1}$ before pruning, we have $\nabla_{\boldsymbol{\widetilde{\mathsf{A}}}}\mathcal{L}=\nabla_{\boldsymbol{\mathsf{A}}}\mathcal{L}\approx\boldsymbol{0}$.
Then we can obtain $\boldsymbol{g}=\nabla_{\boldsymbol{m}}\mathcal{L}\approx\boldsymbol{0}$,
\ie, $g_{i}=\frac{\partial\mathcal{L}}{\partial m_{i}}=\frac{1}{N}\sum\frac{\partial\mathcal{L}_{n}}{\partial m_{i}}\approx0$
where $\mathcal{L}_{n}$ denotes the negative log-likelihood loss
of $n$-th sample and $\frac{\partial\mathcal{L}_{n}}{\partial m_{i}}$
is the sample-wise gradient. We compute the diagonal entry $H_{ii}$
of the Hessian matrix $\boldsymbol{H}$:

\begin{align}
 & H_{ii}=\frac{\partial^{2}\mathcal{L}}{\partial m_{i}^{2}}=\frac{1}{N}\sum_{n=1}^{N}\frac{\partial^{2}\mathcal{L}_{n}}{\partial m_{i}^{2}}\approx-\frac{\partial^{2}}{\partial m_{i}^{2}}\mathbb{E}\left[\log p\left(\boldsymbol{y}\mid\boldsymbol{x}\right)\right]\nonumber \\
 & =\mathbb{E}\left[-\frac{\partial}{\partial m_{i}}\log p\left(\boldsymbol{y}\mid\boldsymbol{x}\right)\right]^{2}\approx\frac{1}{N}\sum_{n=1}^{N}\left(\frac{\partial\mathcal{L}_{n}}{\partial m_{i}}\right)^{2},\label{eq:hessian}
\end{align}
where we employ Fisher information to transform second-order derivative
to the square of first-order derivative. Note that $g_{i}\approx0$
and $H_{ii}\geq0$ which correspond to mean and variance of the sample-wise
gradient, respectively. Assume $\boldsymbol{\mathsf{A}}_{n}\in\mathbb{R}^{c\times h\times w}$
is the feature map of $n$-th example, we have the masked feature
$\boldsymbol{\widetilde{\mathsf{A}}}_{n}=\boldsymbol{\mathsf{A}}_{n}\odot\widetilde{\boldsymbol{m}}$
where $\widetilde{\boldsymbol{m}}\in\mathbb{R}^{c\times h\times w}$
is broadcasted by $\boldsymbol{m}\in\mathbb{R}^{c}$, so we can compute
$\nabla_{\widetilde{\boldsymbol{m}}}\mathcal{L}_{n}=\boldsymbol{\mathsf{A}}_{n}\odot\nabla_{\boldsymbol{\widetilde{\mathsf{A}}}_{n}}\mathcal{L}_{n}\in\mathbb{R}^{c\times h\times w}$,
in which $\nabla_{\boldsymbol{\widetilde{\mathsf{A}}}_{n}}\mathcal{L}_{n}$
is already available during the backward pass without requiring additional
computation.  Then the sample-wise gradient \wrt $\boldsymbol{m}$
can be obtained by summing over the \emph{spatial} dimension $h$
and $w$: $\nabla_{\boldsymbol{m}}\mathcal{L}_{n}=\textrm{sum}(\nabla_{\widetilde{\boldsymbol{m}}}\mathcal{L}_{n})\in\mathbb{R}^{c}$.
Lastly the importance score for a channel can be computed by averaging
the sample-wise gradients:

\begin{equation}
s_{i}=\frac{1}{2N}\sum_{n=1}^{N}\left(\frac{\partial\mathcal{L}_{n}}{\partial m_{i}}\right)^{2}\propto\sum_{n=1}^{N}\left(\frac{\partial\mathcal{L}_{n}}{\partial m_{i}}\right)^{2},
\end{equation}
which is proportional to the squared gradient of the mask. The above
derivation is based on the model convergence, to satisfy it, a greedy
pruning strategy is employed. Starting from a dense model, we first
accumulate the importance scores by passing a few batches, then the
least important channel is pruned. Next we fine-tune the pruned model
and meanwhile re-accumulate the importance scores of the remained
channels, following which the remained least important one is pruned,
and the procedure recurs. 
\begin{figure}
\begin{centering}
\includegraphics[scale=0.7]{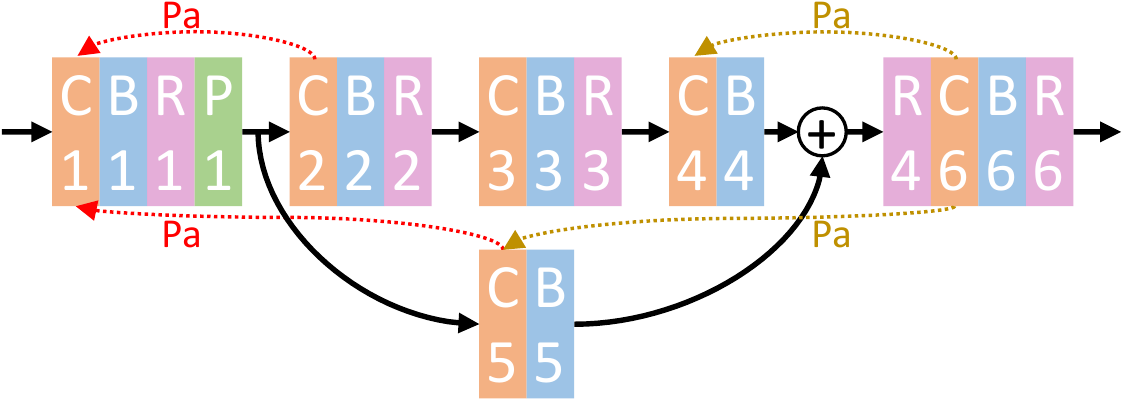}
\par\end{centering}
\caption{\label{fig:dfs}{\small{}Depth-first search for finding parents (Pa).
We take the first residual block of ResNet-50 as an example which
consists of Conv (C), BN (B), ReLU (R) and Pooling (P) layers. During
layer grouping we ignore all layers except Conv layers, so we can
adopt DFS to find that C1 has two children C2/C5, and C6 has two parents
C4/C5. C2 and C5 have coupled channels which should be pruned together
such that output channels of C1 can be pruned.}
}
\end{figure}

\subsection{Prune Coupled Channels}

Till now we can prune early-stage networks like AlexNet \cite{krizhevsky2017imagenet}
and VGGNet \cite{Simonyan15} which involve normal Conv layers and
sequential structures where pruning only affects a layer and its \emph{single}
preceding one. However, recent networks contain complicated structures
such as residual connections \cite{he2016deep}, group convolutions
(GConv) \cite{xie2017aggregated}, depth-wise convolutions (DWConv)
\cite{sandler2018mobilenetv2} and feature pyramid networks (FPN)
\cite{lin2017feature} in object detection. There emerge coupled channels
which should be pruned simultaneously to achieve higher speedup than
pruning only the isolated channels. We propose mask sharing in coupled
channels. Given the network computation graph $\mathcal{G}$ containing
nodes like convolution (Conv), batch normalization (BN), ReLU and
pooling (Pool) layers, we adopt the proposed layer grouping algorithm
to find the coupled channels as Alg. \ref{alg: dfs}. Firstly, we
use depth-first search (DFS) as Fig. \ref{fig:dfs} to find parents
$P\left[l_{i}\right]$ of each Conv/FC layer $l_{i}$. Since channel
pruning only affects the channel dimension, we ignore all layers except
Conv/FC layers in $\mathcal{G}$ during layer grouping. Then given
parents of each layer, we can assign layers to different groups where
layers in one group have coupled channels to be pruned simultaneously.
It contains the following situations: (1) layers which have the same
parents should be assigned to one group because their input channels
(or equivalently, output channels of their parents) are coupled and
should be pruned together as Fig. \ref{fig:prune} (b); (2) layers
whose parents contain GConv should be in the same group with their
parents because the input and output channels of GConv are coupled
as Fig. \ref{fig:prune} (c). For the isolated channels, there is
only one layer in the group such as the $3\times3$ Conv of a residual
bottleneck as Fig. \ref{fig:prune} (a).

After obtaining the coupled channels via layer grouping, we make them
share the same mask. Then the overall contribution of coupled channels
can be computed by:

\begin{equation}
s_{i}\propto\sum_{n=1}^{N}\left(\sum_{x\in\mathbb{X}}\frac{\partial\mathcal{L}_{n}}{\partial m_{i}^{x}}\frac{\partial m_{i}^{x}}{\partial m_{i}}\right)^{2}=\sum_{n=1}^{N}\left(\sum_{x\in\mathbb{X}}\frac{\partial\mathcal{L}_{n}}{\partial m_{i}^{x}}\right)^{2},\label{eq:score}
\end{equation}
where $m_{i}^{x}$ is a copy of $m_{i}$ in channel $x$, one of the
coupled channels in $\mathbb{X}$ that share the same mask $m_{i}$.
The overall importance of coupled channels exactly follows the chain
rule of gradient computation for shared parameters, and thus is principled
without introducing any heuristics. $\mathbb{X}$ can be channels
in the same layer or those distributed in different layers, and thus
summation over $\mathbb{X}$ can be both \emph{in-layer} and \emph{cross-layer}.
For the in-layer case, $\mathbb{X}$ contains channels in one layer
(such as the coupled channels in $i$-th group of a single GConv).
For the cross-layer case (such as residual connections), $\mathbb{X}$
consists of the $i$-th channel from all layers found by our layer
grouping algorithm. 
\begin{figure}
\begin{centering}
\includegraphics[scale=0.45]{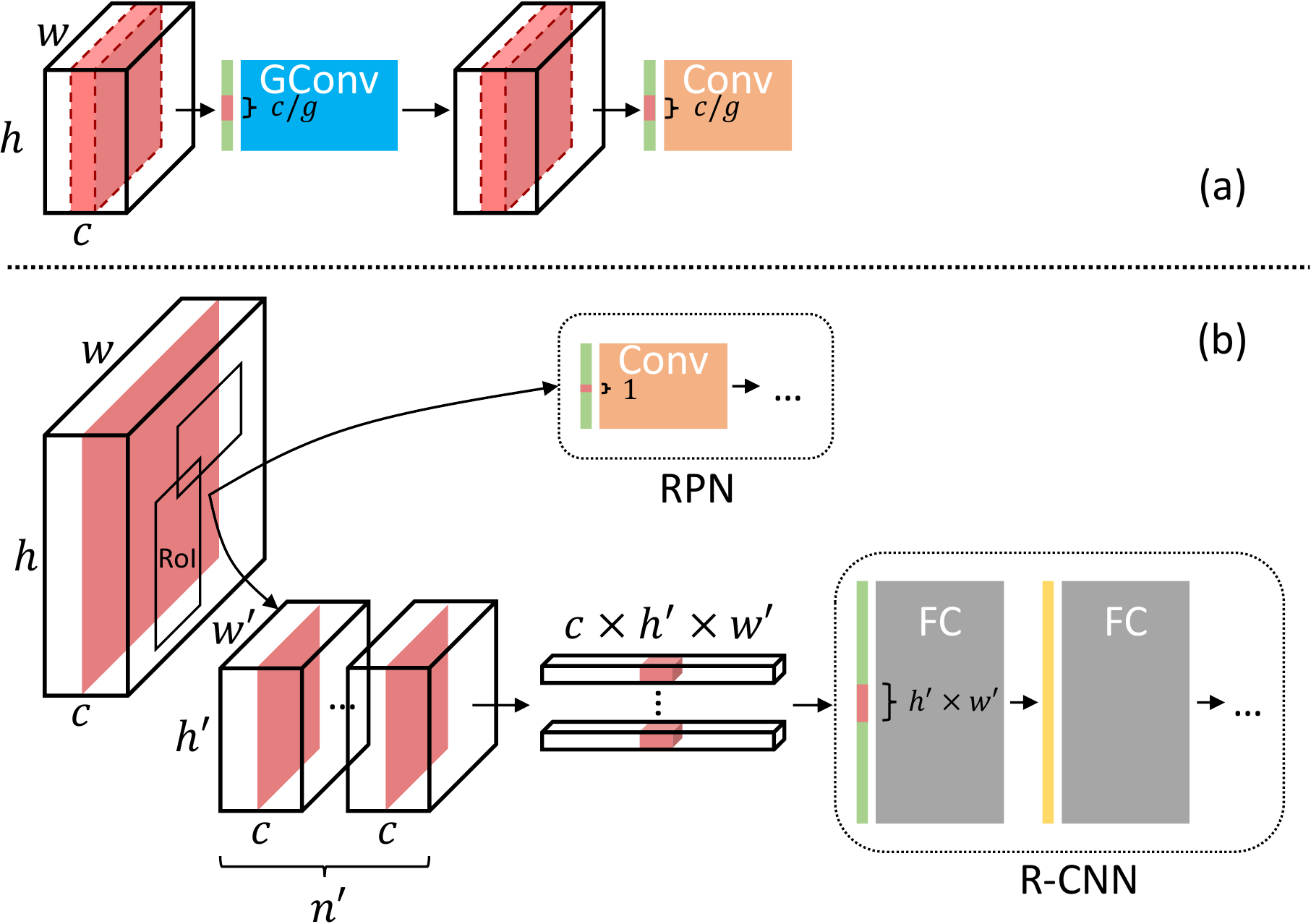}
\par\end{centering}
\caption{\label{fig:conv}{\small{}(a) prune group convolution (GConv). (b)
prune RPN and R-CNN in Faster R-CNN.}
}
\end{figure}

\begin{algorithm}
\textbf{Input:} computational graph $\mathcal{G}$ with layers $\mathbb{L}=\left\{ l_{i}\right\} $

\textbf{Output:} groups of layers $\mathbb{G}=\left\{ g_{i}\right\} $
where $g_{i}=\left\{ l_{j}\right\} $

$\:\:$1: \textbf{for} $l$ \textbf{in} layers $\mathbb{L}$; \textbf{do}

$\:\:$2: $\:\:$find its parents $P\left[l\right]$ via DFS on graph
$\mathcal{G}$

$\:\:$3: \textbf{end for}

$\:\:$4: initialization: groups $\mathbb{G}\leftarrow\emptyset$

$\:\:$5: \textbf{for} $l$ \textbf{in} layers $\mathbb{L}$; \textbf{do}

$\:\:$6: $\:\:$\texttt{new\_group}$\;\leftarrow\;$True

$\:\:$7: $\:\:$\textbf{for} $g$ \textbf{in }groups $\mathbb{G}$;
\textbf{do}

$\:\:$8: $\:\:\:\:$$C\leftarrow\left\{ l^{\prime}\mid l^{\prime}\in g\;\textrm{and}\;l^{\prime}\;\textrm{is}\;\textrm{GConv/DWConv}\right\} $

$\:\:$9: $\:\:\:\:$\textbf{if} $P\left[l\right]\cap\left(P\left[g\right]\cup C\right)\neq\emptyset$;
\textbf{then}

10: $\:\:\:\:\:\:$\textbf{$g\leftarrow g\cup\left\{ l\right\} $},
$P\left[g\right]\leftarrow P\left[g\right]\cup P\left[l\right]$

11: $\:\:\:\:\:\:$\texttt{new\_group}$\;\leftarrow\;$False; \textbf{break}

12: $\:\:$\textbf{end for}

13: $\:\:$\textbf{if} \texttt{new\_group}; \textbf{then}

14: $\:\:\:\:$$g^{\prime}\leftarrow\left\{ l\right\} $, $P\left[g^{\prime}\right]\leftarrow P\left[l\right]$,
$\mathbb{G}\leftarrow\mathbb{G}\cup\left\{ g^{\prime}\right\} $

15: \textbf{end for}

\caption{\label{alg: dfs} Layer Grouping}
\end{algorithm}
\begin{algorithm}
\textbf{Input:} unpruned model $\boldsymbol{W}_{0}$, prune interval
$d$, training data

\textbf{Output:} pruned model $\boldsymbol{W}$, channel masks $\boldsymbol{m}$

$\:\:$1: initialization: $\boldsymbol{W}\leftarrow\boldsymbol{W}_{0}$,
$\boldsymbol{m}\leftarrow\boldsymbol{1}$, $t\leftarrow0$

$\:\:$2: find layers having coupled channels via Alg. \ref{alg: dfs}

$\:\:$3: \textbf{repeat}

$\:\:$4: $\:\:$forward: compute loss $\mathcal{L}$

$\:\:$5: $\:\:$backward: compute gradients of parameters $\boldsymbol{W}$
and

$\:\:\;\;\;\;\:\:$accumulate \emph{memory-normalized} importance
scores

$\:\:$6: $\:\:$update model parameters $\boldsymbol{W}$ by gradient
descent

$\:\:$7: $\:\:$update iteration index $t\leftarrow t+1$

$\:\:$8: $\:\:$\textbf{if} $t\%d=0$; \textbf{then}

$\:\:$9: $\:\:\:\:$prune the least important channel $i$ by setting
$m_{i}=0$

10: $\:\:\:\:$zeroize accumulated importance scores

11: \textbf{until} remained FLOPs reduces to desired amount

\caption{\label{alg: update} Group Fisher Pruning}
\end{algorithm}

For pruning \textbf{GConv} with $c$ input channels divided into $g$
groups as Fig. \ref{fig:prune} (c), each time we prune one \emph{group
of channels} which share the same mask, since generally they represent
related features. We first compute the $c$-dim gradient corresponding
to $c$\emph{ individual} channels and reshape it to $g\times\frac{c}{g}$
(see Fig. \ref{fig:conv} (a)), and do \emph{in-layer} sum over the
last dimension to obtain the $g$-dim gradient corresponding to $g$
\emph{groups} of channels. Next we compute the \emph{cross-layer}
summation across layers: the GConv layer itself and layers which are
in the same group with the GConv including its children layers. Then
we obtain the overall importance with the squared gradients as Eq.
(\ref{eq:score}).

For pruning Faster R-CNN, where the first \textbf{Conv} layer of RPN
and the first \textbf{FC} layer of R-CNN are assigned to the same
group as Fig. \ref{fig:prune} (e), similar computation can be done.
Assume the RoIAlign layer is applied to the image-level feature map
$\boldsymbol{\mathsf{A}}_{n}\in\mathbb{R}^{c\times h\times w}$ and
produces RoI features $\boldsymbol{\mathsf{F}}\in\mathbb{R}^{n^{\prime}\times c\times h^{\prime}\times w^{\prime}}$
from $n^{\prime}$ RoIs in each image (see Fig. \ref{fig:conv} (b)).
Recall that in Faster R-CNN the RoI features will be flattened as
$\boldsymbol{\mathsf{F}}^{\prime}\in\mathbb{R}^{n^{\prime}\times\left(c\times h^{\prime}\times w^{\prime}\right)}$
and sent to the first FC layer of R-CNN. For FC layer we can compute
gradients of its masks with shape of $n^{\prime}\times\left(c\times h^{\prime}\times w^{\prime}\right)$,
which is then transposed to $c\times\left(n^{\prime}\times h^{\prime}\times w^{\prime}\right)$
and \emph{in-layer} summed over the last dimension to obtain the $c$-dim
gradient. Now it has the same shape as gradients computed from the
first Conv layer in RPN. The overall gradients of coupled channels
from the RPN Conv layer and the R-CNN FC layer can be obtained via
\emph{cross-layer} summation. Lastly the gradients are used to compute
the importances as Eq. (\ref{eq:score}).  

\subsection{Importance Normalization}

The raw importance scores do not take into consideration the computation
costs of different channels, however it is more effective to prune
the least important channel with the highest cost. Otherwise we may
prune too many channels to achieve the desired speedup, but it may
lead to degraded accuracy resulting from less parameters retained.
We propose to normalize the importance scores by the computation reduction
of each channel. We first try the widely-used FLOPs proxy and normalize
the importance by the reduction of FLOPs $\triangle C$/$\triangle C_{g}$
for pruning an \emph{input} channel of normal Conv/GConv as $\triangle C=n\times c_{o}^{\prime}\times h\times w\times k_{h}\times k_{w}$
and $\triangle C_{g}=n\times\frac{c_{o}}{g}\times h\times w\times\frac{c_{i}}{g}\times k_{h}\times k_{w}$,
where $k_{h}/k_{w}$ denotes the kernel height/width and $g$ is the
group number in GConv. Different from previous methods which compute
channel FLOPs in advance and fix them during pruning, we dynamically
update the FLOPs by remained channels of the network. We use $c_{i}^{\prime}\leq c_{i}$
and $c_{o}^{\prime}\leq c_{o}$ to represent the unpruned input and
output channel number of each layer. As the layers are connected internally,
pruning a channel not only brings FLOPs reduction in the current layer,
but also in its parent layers across the computation graph, which
can be computed as $\triangle C^{p}=n\times h\times w\times c_{i}^{\prime}\times k_{h}\times k_{w}$
and $\triangle C_{g}^{p}=n\times\frac{c_{o}}{g}\times h\times w\times c_{i}^{\prime}\times k_{h}\times k_{w}.$

\begin{figure}
\begin{centering}
\subfloat[{\small{}Normalize by FLOPs.}]{\begin{centering}
\includegraphics[scale=0.18]{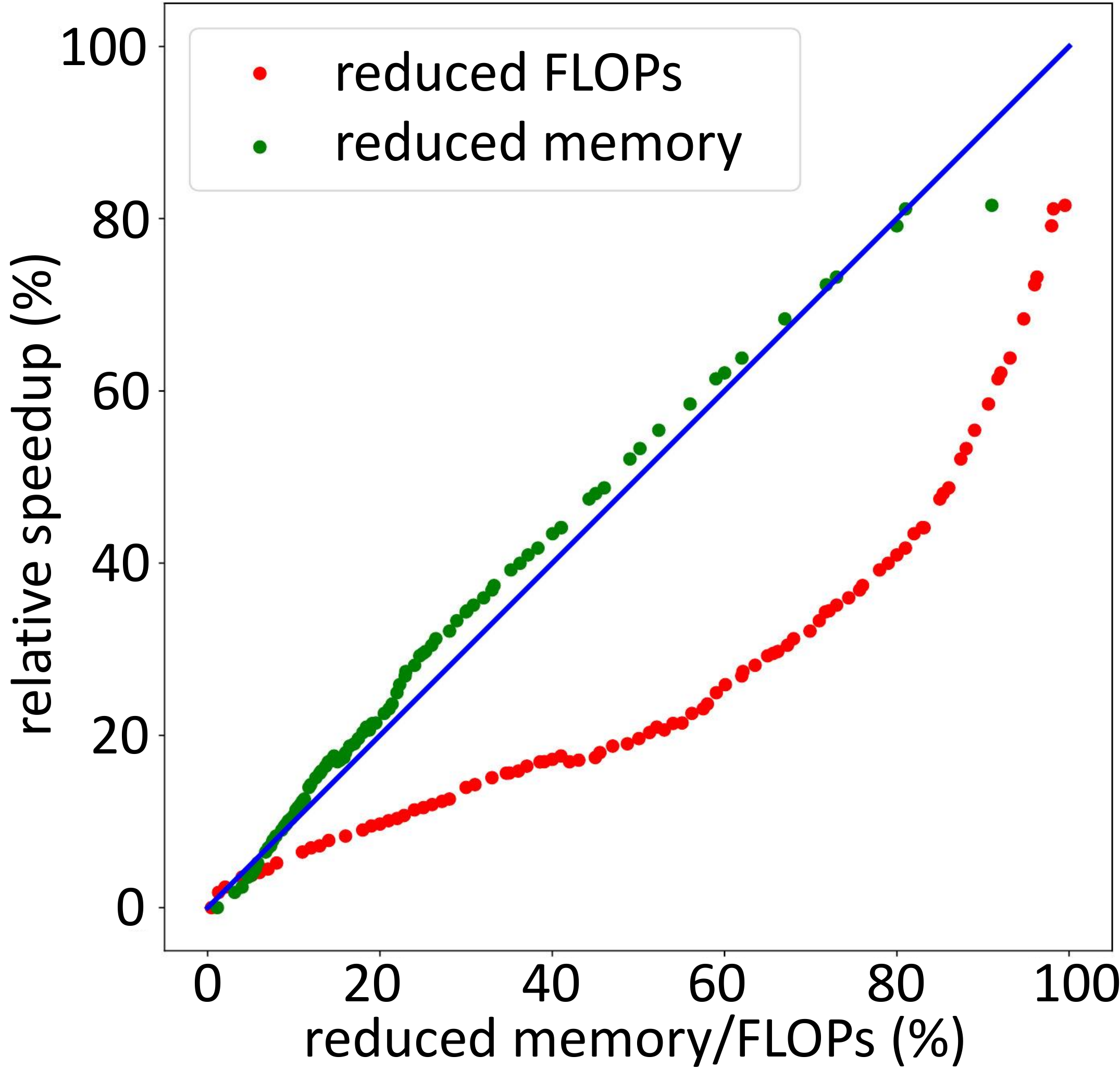}
\par\end{centering}
}\subfloat[{\small{}Normalize by memory.}]{\begin{centering}
\includegraphics[scale=0.18]{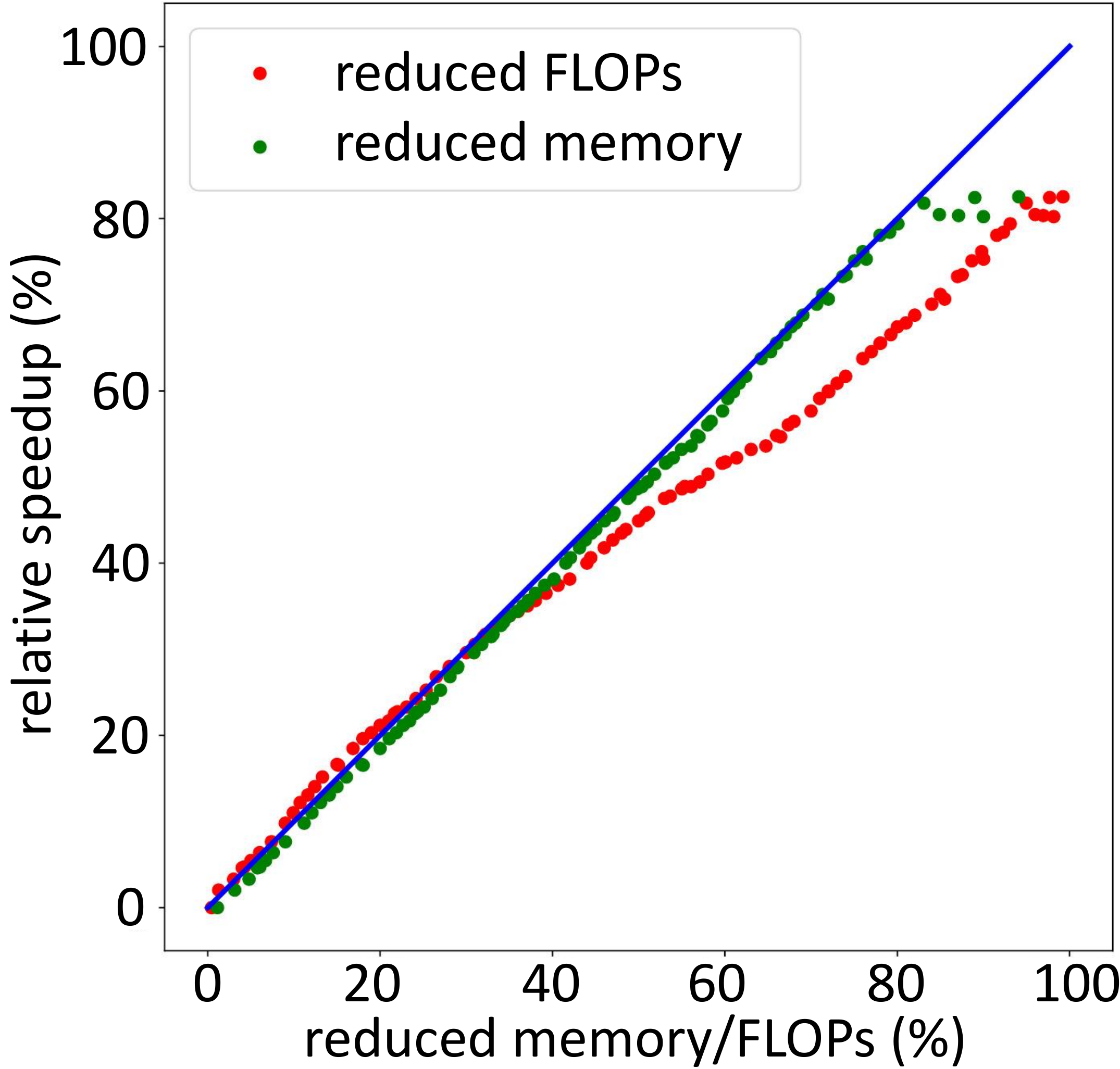}
\par\end{centering}
}
\par\end{centering}

\caption{\label{fig:norm}{\small{}The comparison between normalizing importances
by FLOPs and memory. States of the model (reduced FLOPs/memory and
relative speedup) during the pruning process are shown.}
}
\end{figure}

However, we find that reduction of FLOPs is not directly correlated
with the inference speedup. In contrast, reduction of memory increases
linearly with speedup (Fig. \ref{fig:norm} (a)), which motivates
us to employ the memory reduction as the importance normalization.
The memory reduction of pruning one channel can be obtained for normal
Conv and GConv as $\triangle M=n\times h\times w$ and $\triangle M_{g}=n\times\frac{c}{g}\times h\times w$.
Note that we discard one group at a time when pruning GConv. Similar
to FLOPs reduction, pruning an input channel in one layer brings memory
reduction from all of its parents (which can be found by DFS as Fig.
\ref{fig:dfs}), and we obtain the overall reduction by summing separate
values. Finally we adopt the memory-normalized importance $s_{i}/\triangle M$
of each channel to evaluate its significance and prune the least important
one every few iterations. There exist methods \cite{wang2020hat,li2020benanza}
that directly profile the running time without resorting to proxies.
However, it is not applicable here since we prune in a fine-grained
manner. The running time difference of discarding one or few channels
is too subtle to measure. Through memory-normalization (Fig. \ref{fig:norm}
(b)), we notice that in the pruned model, the reduction of memory
is still linearly correlated with speedup. Besides, the reduction
of FLOPs is more correlated with speedup than the FLOPs-normalization
(Fig. \ref{fig:norm} (a)) variant. In the appendix we demonstrate
that the memory is a good proxy generally applicable to various networks.

\section{Experiments}

In this section we first conduct ablation studies to verify the effectiveness
of our layer grouping and mask sharing strategy to prune coupled channels,
and that of the proposed memory normalized importance scores. We measure
the batch inference time on NVIDIA 2080 Ti GPU to prove our pruned
models can significantly accelerate inference with little accuracy
drop. Next we show that our method can outperform previous methods
to prune various networks under different FLOPs constraints including
the rather compact ResNet \cite{he2016deep} and ResNeXt \cite{xie2017aggregated}
where residual connection and GConv is adopted. Our method can be
applied to prune MobileNetV2 \cite{sandler2018mobilenetv2} where
DWConv is presented, and it outperforms the uniform-scaled baselines
remarkably. It can also be used to prune RegNet \cite{radosavovic2020designing}
which is neural architecture search based and highly efficient and
accurate, surprisingly we achieve higher accuracy and speed than the
searched counterpart under the same FLOPs. Finally, we prune object
detection networks with sophisticated structures and show significant
speedup with negligible mAP drop.  We conduct all experiments for
the task of image classification and object detection on the ImageNet
\cite{deng2009imagenet} and COCO \cite{lin2014microsoft} datasets,
respectively. We prune a channel every $d=25/10$ iterations when
pruning classification/detection networks. After the whole pruning
process we fine-tune the pruned model for the same number of epochs
that is used to train the unpruned model, which is trained following
standard practices. The complete pruning pipeline of our proposed
method is in Alg. \ref{alg: update}. All experiments are done using
PyTorch \cite{paszke2017automatic} and more details can be found
in the appendix.

\subsection{Ablation Studies}

As shown in Tab. \ref{tab:group}, pruning coupled layers simultaneously
(``-M'') as Fig. \ref{fig:prune} (b) rather than the isolated channels
only (``-I'') as Fig. \ref{fig:prune} (a), for residual networks
with both 50 and 101 layers, we achieve higher or comparable top-1
accuracy but with much higher inference speed, which verifies we can
obtain better actual speedup under the same FLOPs. We also adopt the
absolute value of the first-order gradient as the importance metric,
but for pruning ResNet-50 it only obtains 75.8\% top-1 accuracy, which
lags behind our method based on Fisher information 76.4\%.

Besides, we explore different importance normalization strategies:
unnormalized (``-U''), normalized by FLOPs reduction (``-F'')
and normalized by memory reduction (``-M'') in Fig. \ref{fig:regnet}
(c) and Tab. \ref{tab:group}. We find that normalizing importance
scores by memory reduction can achieve the best accuracy-efficiency
trade-off compared with the other two variants. The unnormalized importance
score brings the worst efficiency gain and the largest accuracy drop,
which results from the least parameter remained. For ResNet, ResNeXt
and MobileNet, normalization by memory reduction is more efficiency-friendly
and with higher accuracy. 
\begin{table}
\caption{\label{tab:group}{\small{}Prune ResNet, ResNeXt and MobileNetV2 on
ImageNet. The }\textbf{\small{}column}{\small{} ``T1'' represents
top-1 accuracy on the validation set, ``F'' denotes FLOPs, ``P''
is number of parameters, ``M'' means memory, ``T'' is the inference
time and ``S'' shows the speedup on GPUs. The }\textbf{\small{}row}{\small{}
with no suffix represents the unpruned model, ``-I'' shows the results
of pruning only the internal layers, ``-M'' is our full method which
normalizes importance scores by memory reduction and prunes coupled
channels via layer grouping, ``-F'' normalizes the importance scores
with FLOPs reduction and ``-U'' uses the raw Fisher information
without normalization.}}

\begin{spacing}{0.7}
\centering{}{\scriptsize{}}%
\begin{tabular}{l|llllll}
{\scriptsize{}model} & {\scriptsize{}T1(\%)} & {\scriptsize{}F($10^{9}$)} & {\scriptsize{}P($10^{6}$)} & {\scriptsize{}M($10^{6}$)} & {\scriptsize{}T(ms)} & {\scriptsize{}S($\times$)}\tabularnewline
\hline 
{\scriptsize{}Res50} & {\scriptsize{}76.79} & {\scriptsize{}4.089} & {\scriptsize{}25.55} & {\scriptsize{}11.11} & {\scriptsize{}65.48} & {\scriptsize{}-}\tabularnewline
{\scriptsize{}Res50-I} & {\scriptsize{}76.23} & {\scriptsize{}2.044} & {\scriptsize{}19.96} & {\scriptsize{}9.24} & {\scriptsize{}50.22} & {\scriptsize{}1.30}\tabularnewline
{\scriptsize{}Res50-M} & {\scriptsize{}76.42} & {\scriptsize{}2.044} & {\scriptsize{}19.42} & {\scriptsize{}5.82} & {\scriptsize{}36.50} & {\scriptsize{}1.79}\tabularnewline
{\scriptsize{}Res50-F} & {\scriptsize{}76.20} & {\scriptsize{}2.043} & {\scriptsize{}17.17} & {\scriptsize{}8.27} & {\scriptsize{}47.07} & {\scriptsize{}1.39}\tabularnewline
{\scriptsize{}Res50-U} & {\scriptsize{}74.93} & {\scriptsize{}2.044} & {\scriptsize{}8.881} & {\scriptsize{}9.27} & {\scriptsize{}50.06} & {\scriptsize{}1.31}\tabularnewline
\hline 
{\scriptsize{}Res101} & {\scriptsize{}78.29} & {\scriptsize{}7.801} & {\scriptsize{}44.54} & {\scriptsize{}16.23} & {\scriptsize{}106.7} & {\scriptsize{}-}\tabularnewline
{\scriptsize{}Res101-I} & {\scriptsize{}78.36} & {\scriptsize{}3.900} & {\scriptsize{}28.04} & {\scriptsize{}14.14} & {\scriptsize{}84.93} & {\scriptsize{}1.26}\tabularnewline
{\scriptsize{}Res101-M} & {\scriptsize{}78.33} & {\scriptsize{}3.900} & {\scriptsize{}28.02} & {\scriptsize{}10.84} & {\scriptsize{}71.20} & {\scriptsize{}1.50}\tabularnewline
{\scriptsize{}Res101-F} & {\scriptsize{}78.30} & {\scriptsize{}3.899} & {\scriptsize{}26.89} & {\scriptsize{}14.13} & {\scriptsize{}83.91} & {\scriptsize{}1.27}\tabularnewline
{\scriptsize{}Res101-U} & {\scriptsize{}78.14} & {\scriptsize{}3.900} & {\scriptsize{}24.25} & {\scriptsize{}14.49} & {\scriptsize{}84.10} & {\scriptsize{}1.27}\tabularnewline
\hline 
{\scriptsize{}NeXt50} & {\scriptsize{}77.97} & {\scriptsize{}4.230} & {\scriptsize{}25.02} & {\scriptsize{}14.40} & {\scriptsize{}87.16} & {\scriptsize{}-}\tabularnewline
{\scriptsize{}NeXt50-M} & {\scriptsize{}77.53} & {\scriptsize{}2.115} & {\scriptsize{}18.05} & {\scriptsize{}9.02} & {\scriptsize{}55.35} & {\scriptsize{}1.57}\tabularnewline
{\scriptsize{}NeXt50-F} & {\scriptsize{}77.50} & {\scriptsize{}2.114} & {\scriptsize{}13.10} & {\scriptsize{}10.57} & {\scriptsize{}59.79} & {\scriptsize{}1.46}\tabularnewline
{\scriptsize{}NeXt50-U} & {\scriptsize{}76.97} & {\scriptsize{}2.113} & {\scriptsize{}8.426} & {\scriptsize{}11.17} & {\scriptsize{}60.23} & {\scriptsize{}1.45}\tabularnewline
\hline 
{\scriptsize{}MBv2} & {\scriptsize{}77.41} & {\scriptsize{}1.137} & {\scriptsize{}11.25} & {\scriptsize{}13.35} & {\scriptsize{}44.21} & {\scriptsize{}-}\tabularnewline
{\scriptsize{}MBv2-M} & {\scriptsize{}75.97} & {\scriptsize{}0.568} & {\scriptsize{}6.05} & {\scriptsize{}6.84} & {\scriptsize{}24.72} & {\scriptsize{}1.79}\tabularnewline
{\scriptsize{}MBv2-F} & {\scriptsize{}75.97} & {\scriptsize{}0.566} & {\scriptsize{}5.31} & {\scriptsize{}11.08} & {\scriptsize{}37.58} & {\scriptsize{}1.18}\tabularnewline
{\scriptsize{}MBv2-U} & {\scriptsize{}72.94} & {\scriptsize{}0.569} & {\scriptsize{}2.27} & {\scriptsize{}9.819} & {\scriptsize{}31.78} & {\scriptsize{}1.39}\tabularnewline
\end{tabular}{\scriptsize\par}
\end{spacing}
\end{table}

Except for the rather compact residual networks, we also prune the
light-weight networks MobileNetV2 (MBv2). In Fig. \ref{fig:regnet}
(b) and Tab. \ref{tab:mb} we compare the accuracy and speed of our
pruned networks and the uniform scaled ones under different FLOPs
budgets. To obtain a network with similar FLOPs as MBv2 (\eg, MBv2-0.7$\times$),
we prune a uniform-scaled double-FLOPs MBv2 (\eg, MBv2-1.0$\times$)
to 50\% FLOPs remained. It can be seen that our pruned networks significantly
outperform the uniform-scaled baselines. Other than only pruning the
human-designed networks, we prune the highly-efficient RegNet to show
that we can prune it to further boost the efficiency and accuracy.
We prune RegNet (\eg, RegX-1.6G) to 50\% FLOPs remained to compare
with the searched half-FLOPs RegNet (\eg, RegX-0.8G). As shown in
Fig. \ref{fig:regnet} (a) and Tab. \ref{tab:reg}, our pruned networks
outperform the searched counterparts in this extreme circumstance.

\begin{table}
\caption{\label{tab:mb}{\small{}Prune MobileNetV2 on ImageNet.}}

\begin{spacing}{0.7}
\begin{centering}
{\scriptsize{}}%
\begin{tabular}{c|ccccc}
{\scriptsize{}model} & {\scriptsize{}T1 (\%)} & {\scriptsize{}F ($10^{9}$)} & {\scriptsize{}P ($10^{6}$)} & {\scriptsize{}M ($10^{6}$)} & {\scriptsize{}T (ms)}\tabularnewline
\hline 
{\scriptsize{}MBv2-2.0$\times$} & {\scriptsize{}77.41} & {\scriptsize{}1.14} & {\scriptsize{}11.25} & {\scriptsize{}13.35} & {\scriptsize{}44.21}\tabularnewline
{\scriptsize{}Ours} & {\scriptsize{}77.00} & {\scriptsize{}1.09} & {\scriptsize{}11.06} & {\scriptsize{}9.66} & {\scriptsize{}37.22}\tabularnewline
\hline 
{\scriptsize{}MBv2-1.4$\times$} & {\scriptsize{}75.74} & {\scriptsize{}0.58} & {\scriptsize{}6.11} & {\scriptsize{}9.57} & {\scriptsize{}30.63}\tabularnewline
{\scriptsize{}Ours} & {\scriptsize{}75.97} & {\scriptsize{}0.57} & {\scriptsize{}6.05} & {\scriptsize{}6.84} & {\scriptsize{}24.72}\tabularnewline
\hline 
{\scriptsize{}MBv2-1.0$\times$} & {\scriptsize{}72.92} & {\scriptsize{}0.30} & {\scriptsize{}3.50} & {\scriptsize{}6.68} & {\scriptsize{}20.82}\tabularnewline
{\scriptsize{}Ours} & {\scriptsize{}73.42} & {\scriptsize{}0.29} & {\scriptsize{}3.31} & {\scriptsize{}4.82} & {\scriptsize{}16.61}\tabularnewline
\hline 
{\scriptsize{}MBv2-0.7$\times$} & {\scriptsize{}68.98} & {\scriptsize{}0.17} & {\scriptsize{}2.48} & {\scriptsize{}5.26} & {\scriptsize{}15.81}\tabularnewline
{\scriptsize{}Ours} & {\scriptsize{}69.16} & {\scriptsize{}0.15} & {\scriptsize{}1.81} & {\scriptsize{}3.39} & {\scriptsize{}11.62}\tabularnewline
\hline 
{\scriptsize{}MBv2-0.5$\times$} & {\scriptsize{}64.70} & {\scriptsize{}0.10} & {\scriptsize{}1.97} & {\scriptsize{}3.64} & {\scriptsize{}10.97}\tabularnewline
{\scriptsize{}Ours} & {\scriptsize{}64.68} & {\scriptsize{}0.09} & {\scriptsize{}1.17} & {\scriptsize{}2.59} & {\scriptsize{}8.51}\tabularnewline
\end{tabular}{\scriptsize\par}
\par\end{centering}
\end{spacing}
\end{table}
\begin{table}
\caption{\label{tab:reg}{\small{}Prune RegNetX on ImageNet.}}

\begin{spacing}{0.7}
\centering{}{\scriptsize{}}%
\begin{tabular}{c|ccccc}
{\scriptsize{}model} & {\scriptsize{}T1 (\%)} & {\scriptsize{}F ($10^{9}$)} & {\scriptsize{}P ($10^{6}$)} & {\scriptsize{}M ($10^{6}$)} & {\scriptsize{}T (ms)}\tabularnewline
\hline 
{\scriptsize{}RegX-3.2G} & {\scriptsize{}78.72} & {\scriptsize{}3.176} & {\scriptsize{}15.29} & {\scriptsize{}11.36} & {\scriptsize{}65.99}\tabularnewline
{\scriptsize{}Ours} & {\scriptsize{}78.80} & {\scriptsize{}3.228} & {\scriptsize{}14.34} & {\scriptsize{}9.89} & {\scriptsize{}61.34}\tabularnewline
\hline 
{\scriptsize{}RegX-1.6G} & {\scriptsize{}77.29} & {\scriptsize{}1.602} & {\scriptsize{}9.19} & {\scriptsize{}7.93} & {\scriptsize{}41.21}\tabularnewline
{\scriptsize{}Ours} & {\scriptsize{}77.97} & {\scriptsize{}1.588} & {\scriptsize{}9.30} & {\scriptsize{}7.29} & {\scriptsize{}40.44}\tabularnewline
\hline 
{\scriptsize{}RegX-0.8G} & {\scriptsize{}75.24} & {\scriptsize{}0.799} & {\scriptsize{}7.26} & {\scriptsize{}5.15} & {\scriptsize{}26.71}\tabularnewline
{\scriptsize{}Ours} & {\scriptsize{}76.39} & {\scriptsize{}0.799} & {\scriptsize{}5.93} & {\scriptsize{}5.38} & {\scriptsize{}25.51}\tabularnewline
\hline 
{\scriptsize{}RegX-0.4G} & {\scriptsize{}72.28} & {\scriptsize{}0.398} & {\scriptsize{}5.16} & {\scriptsize{}3.14} & {\scriptsize{}16.32}\tabularnewline
{\scriptsize{}Ours} & {\scriptsize{}73.75} & {\scriptsize{}0.399} & {\scriptsize{}5.42} & {\scriptsize{}2.78} & {\scriptsize{}14.97}\tabularnewline
\hline 
{\scriptsize{}RegX-0.2G} & {\scriptsize{}68.65} & {\scriptsize{}0.199} & {\scriptsize{}2.68} & {\scriptsize{}2.16} & {\scriptsize{}10.09}\tabularnewline
{\scriptsize{}Ours} & {\scriptsize{}70.01} & {\scriptsize{}0.199} & {\scriptsize{}2.77} & {\scriptsize{}2.13} & {\scriptsize{}11.49}\tabularnewline
\end{tabular}{\scriptsize\par}
\end{spacing}
\end{table}

\begin{table}
\caption{\label{tab:sota}{\small{}Compare with SoTAs on ImageNet. The }\textbf{\small{}column}{\small{}
``T1'' represents top-1 accuracy of the pruned model on the validation
set where $\downarrow$ shows the accuracy drop compared with the
unpruned model. ``B1'' shows the top-1 accuracy of the unpruned
base model. ``F'' shows the amount of FLOPs of the pruned model,
where $\downarrow$ elements show the relative FLOPs reduction compared
with the unpruned model. ``S'' denotes the actual speedup of the
pruned model on GPUs.}}

\begin{spacing}{0.7}
\begin{centering}
{\scriptsize{}}%
\begin{tabular}{c|l|ll|ll}
 & {\scriptsize{}method} & {\scriptsize{}T1(\%)} & {\scriptsize{}B1(\%)} & {\scriptsize{}F(G)} & {\scriptsize{}S($\times$)}\tabularnewline
\hline 
\multirow{29}{*}{\begin{turn}{90}
{\scriptsize{}Res50}
\end{turn}} & {\scriptsize{}ThiNet \cite{luo2017thinet}} & {\scriptsize{}74.03} & {\scriptsize{}75.30} & {\scriptsize{}2.58} & {\scriptsize{}1.13}\tabularnewline
 & {\scriptsize{}SSS \cite{huang2018data}} & {\scriptsize{}75.44} & {\scriptsize{}76.12} & {\scriptsize{}3.47} & {\scriptsize{}-}\tabularnewline
 & {\scriptsize{}IE \cite{molchanov2019importance}} & {\scriptsize{}76.43} & {\scriptsize{}76.18} & {\scriptsize{}3.27} & {\scriptsize{}-}\tabularnewline
 & {\scriptsize{}HetConv \cite{singh20het}} & {\scriptsize{}76.16} & {\scriptsize{}76.16} & {\scriptsize{}2.85} & {\scriptsize{}-}\tabularnewline
 & {\scriptsize{}Meta \cite{liu2019metapruning}} & {\scriptsize{}76.20} & {\scriptsize{}76.60} & {\scriptsize{}3.0} & {\scriptsize{}-}\tabularnewline
 & {\scriptsize{}GBN \cite{you2019gate}} & {\scriptsize{}76.19} & {\scriptsize{}75.85} & {\scriptsize{}2.43} & {\scriptsize{}-}\tabularnewline
 & {\scriptsize{}Ours} & {\scriptsize{}76.95} & {\scriptsize{}76.79} & {\scriptsize{}3.06} & {\scriptsize{}1.30}\tabularnewline
\cline{2-6} \cline{3-6} \cline{4-6} \cline{5-6} \cline{6-6} 
 & {\scriptsize{}ThiNet \cite{luo2017thinet}} & {\scriptsize{}72.03} & {\scriptsize{}75.30} & {\scriptsize{}1.81} & {\scriptsize{}1.27}\tabularnewline
 & {\scriptsize{}CP \cite{he2017channel}} & {\scriptsize{}75.06} & {\scriptsize{}76.13} & {\scriptsize{}2.04} & {\scriptsize{}-}\tabularnewline
 & {\scriptsize{}NISP \cite{yu2018nisp}} & {\scriptsize{}0.89$\downarrow$} & {\scriptsize{}-} & {\scriptsize{}2.29} & {\scriptsize{}-}\tabularnewline
 & {\scriptsize{}SFP \cite{he2018soft}} & {\scriptsize{}74.61} & {\scriptsize{}76.15} & {\scriptsize{}2.38} & {\scriptsize{}1.43}\tabularnewline
 & {\scriptsize{}GDP \cite{lin2018accelerating}} & {\scriptsize{}72.61} & {\scriptsize{}75.13} & {\scriptsize{}2.24} & {\scriptsize{}1.24}\tabularnewline
 & {\scriptsize{}SSS \cite{huang2018data}} & {\scriptsize{}71.82} & {\scriptsize{}76.12} & {\scriptsize{}2.33} & {\scriptsize{}-}\tabularnewline
 & {\scriptsize{}DCP \cite{zhuang2018discrimination}} & {\scriptsize{}74.95} & {\scriptsize{}76.01} & {\scriptsize{}1.81} & {\scriptsize{}-}\tabularnewline
 & {\scriptsize{}AOFP \cite{Ding2019ApproximatedOF}} & {\scriptsize{}75.11} & {\scriptsize{}75.34} & {\scriptsize{}1.77} & {\scriptsize{}-}\tabularnewline
 & {\scriptsize{}FPGM \cite{he2019filter}} & {\scriptsize{}74.83} & {\scriptsize{}76.15} & {\scriptsize{}1.90} & {\scriptsize{}1.62}\tabularnewline
 & {\scriptsize{}IE \cite{molchanov2019importance}} & {\scriptsize{}74.50} & {\scriptsize{}76.18} & {\scriptsize{}2.25} & {\scriptsize{}-}\tabularnewline
 & {\scriptsize{}C-SGD \cite{ding2019centripetal}} & {\scriptsize{}74.54} & {\scriptsize{}75.33} & {\scriptsize{}2.20} & {\scriptsize{}-}\tabularnewline
 & {\scriptsize{}Meta \cite{liu2019metapruning}} & {\scriptsize{}75.40} & {\scriptsize{}76.60} & {\scriptsize{}2.0} & {\scriptsize{}-}\tabularnewline
 & {\scriptsize{}GBN \cite{you2019gate}} & {\scriptsize{}75.18} & {\scriptsize{}75.85} & {\scriptsize{}1.84} & {\scriptsize{}-}\tabularnewline
 & {\scriptsize{}LFPC \cite{he2020learning}} & {\scriptsize{}74.46} & {\scriptsize{}76.15} & {\scriptsize{}1.60} & {\scriptsize{}-}\tabularnewline
 & {\scriptsize{}HRank \cite{lin2020hrank}} & {\scriptsize{}74.98} & {\scriptsize{}76.15} & {\scriptsize{}2.30} & {\scriptsize{}-}\tabularnewline
 & {\scriptsize{}Ours} & {\scriptsize{}76.42} & {\scriptsize{}76.79} & {\scriptsize{}2.04} & {\scriptsize{}1.79}\tabularnewline
\cline{2-6} \cline{3-6} \cline{4-6} \cline{5-6} \cline{6-6} 
 & {\scriptsize{}ThiNet \cite{luo2017thinet}} & {\scriptsize{}68.17} & {\scriptsize{}75.30} & {\scriptsize{}1.17} & {\scriptsize{}1.35}\tabularnewline
 & {\scriptsize{}GDP \cite{lin2018accelerating}} & {\scriptsize{}70.93} & {\scriptsize{}75.13} & {\scriptsize{}1.57} & {\scriptsize{}-}\tabularnewline
 & {\scriptsize{}IE \cite{molchanov2019importance}} & {\scriptsize{}71.69} & {\scriptsize{}76.18} & {\scriptsize{}1.34} & {\scriptsize{}-}\tabularnewline
 & {\scriptsize{}Meta \cite{liu2019metapruning}} & {\scriptsize{}73.40} & {\scriptsize{}76.60} & {\scriptsize{}1.0} & {\scriptsize{}-}\tabularnewline
 & {\scriptsize{}CURL \cite{luo2020neural}} & {\scriptsize{}73.39} & {\scriptsize{}76.15} & {\scriptsize{}1.11} & {\scriptsize{}-}\tabularnewline
 & {\scriptsize{}Ours} & {\scriptsize{}73.94} & {\scriptsize{}76.79} & {\scriptsize{}1.02} & {\scriptsize{}2.94}\tabularnewline
\hline 
\multirow{7}{*}{\begin{turn}{90}
{\scriptsize{}Res101}
\end{turn}} & {\scriptsize{}ISTA \cite{ye2018rethinking}} & {\scriptsize{}75.27} & {\scriptsize{}76.40} & {\scriptsize{}4.47} & {\scriptsize{}-}\tabularnewline
 & {\scriptsize{}SFP \cite{he2018soft}} & {\scriptsize{}77.51} & {\scriptsize{}77.37} & {\scriptsize{}4.51} & {\scriptsize{}-}\tabularnewline
 & {\scriptsize{}SSS \cite{huang2018data}} & {\scriptsize{}75.44} & {\scriptsize{}76.40} & {\scriptsize{}3.47} & {\scriptsize{}-}\tabularnewline
 & {\scriptsize{}AOFP \cite{Ding2019ApproximatedOF}} & {\scriptsize{}76.40} & {\scriptsize{}76.63} & {\scriptsize{}3.89} & {\scriptsize{}-}\tabularnewline
 & {\scriptsize{}FPGM \cite{he2019filter}} & {\scriptsize{}77.32} & {\scriptsize{}77.37} & {\scriptsize{}4.51} & {\scriptsize{}-}\tabularnewline
 & {\scriptsize{}IE \cite{molchanov2019importance}} & {\scriptsize{}77.35} & {\scriptsize{}77.37} & {\scriptsize{}4.70} & {\scriptsize{}-}\tabularnewline
 & {\scriptsize{}Ours} & {\scriptsize{}78.33} & {\scriptsize{}78.29} & {\scriptsize{}3.90} & {\scriptsize{}1.50}\tabularnewline
\hline 
\multirow{7}{*}{\begin{turn}{90}
{\scriptsize{}MBv2}
\end{turn}} & {\scriptsize{}AMC \cite{he2018amc}} & {\scriptsize{}70.80} & {\scriptsize{}71.80} & {\scriptsize{}0.22} & {\scriptsize{}-}\tabularnewline
 & {\scriptsize{}Meta \cite{liu2019metapruning}} & {\scriptsize{}72.70} & {\scriptsize{}74.70} & {\scriptsize{}0.29} & {\scriptsize{}-}\tabularnewline
 & {\scriptsize{}Ours} & {\scriptsize{}73.42} & {\scriptsize{}75.74} & {\scriptsize{}0.29} & {\scriptsize{}1.84}\tabularnewline
\cline{2-6} \cline{3-6} \cline{4-6} \cline{5-6} \cline{6-6} 
 & {\scriptsize{}Meta \cite{liu2019metapruning}} & {\scriptsize{}68.20} & {\scriptsize{}74.70} & {\scriptsize{}0.14} & {\scriptsize{}-}\tabularnewline
 & {\scriptsize{}Ours} & {\scriptsize{}69.16} & {\scriptsize{}75.74} & {\scriptsize{}0.15} & {\scriptsize{}1.79}\tabularnewline
\cline{2-6} \cline{3-6} \cline{4-6} \cline{5-6} \cline{6-6} 
 & {\scriptsize{}Meta \cite{liu2019metapruning}} & {\scriptsize{}65.00} & {\scriptsize{}74.70} & {\scriptsize{}0.11} & {\scriptsize{}-}\tabularnewline
 & {\scriptsize{}Ours} & {\scriptsize{}65.94} & {\scriptsize{}75.74} & {\scriptsize{}0.10} & {\scriptsize{}1.82}\tabularnewline
\hline 
\multirow{3}{*}{\begin{turn}{90}
{\scriptsize{}NeXt50}
\end{turn}} &  &  &  &  & \tabularnewline
 & {\scriptsize{}SSS \cite{huang2018data}} & {\scriptsize{}74.98} & {\scriptsize{}77.57} & {\scriptsize{}2.43} & {\scriptsize{}-}\tabularnewline
 & {\scriptsize{}Ours} & {\scriptsize{}77.53} & {\scriptsize{}77.97} & {\scriptsize{}2.11} & {\scriptsize{}1.57}\tabularnewline
\end{tabular}{\scriptsize\par}
\par\end{centering}
\end{spacing}
\end{table}

\subsection{Compare with SoTAs}

To compare with previous state-of-the-arts, we conduct extensive experiments
of image classification on ImageNet using different network structures
and FLOPs constraints. From Fig. \ref{fig:sota} and Tab. \ref{tab:sota}
we can see that our method performs best. Specifically, we outperform
layer-wise pruning methods such as CP \cite{he2017channel} and ThiNet
\cite{luo2017thinet} because we evaluate the importance scores globally
throughout the network. Moreover, we do not need sensitivity analysis
which is required by NISP \cite{yu2018nisp} to decide the pruning
ratio for each layer, as our method can automatically learn to prune
the least important channels considering the current state of the
network. We also achieve better accuracy than the methods C-SGD \cite{ding2019centripetal},
GBN \cite{you2019gate} and IE \cite{molchanov2019importance} which
compute the overall importance of coupled channels via heuristics,
validating the benefits of our importance metric grounded on gradients
obtained by the principled chain rule. 
\begin{table}
\caption{\label{tab:det}{\small{}Prune detection networks including RetinaNet,
FSAF, ATSS, PAA and Faster R-CNN on COCO. The ``AP'' column shows
the mAP (\%) on the validation set and the ``T'' column is the inference
time. The other columns and the suffix in rows have the same meaning
as Tab. \ref{tab:group} ($^{\star}:$ no Group Normalization in heads
to compare with RetinaNet).}}

\begin{spacing}{0.7}
\centering{}{\scriptsize{}}%
\begin{tabular}{l|llllll}
{\scriptsize{}model} & {\scriptsize{}AP} & {\scriptsize{}F($10^{9}$)} & {\scriptsize{}P($10^{6}$)} & {\scriptsize{}M($10^{6}$)} & {\scriptsize{}T(ms)} & {\scriptsize{}S($\times$)}\tabularnewline
\hline 
{\scriptsize{}Retina} & {\scriptsize{}36.5} & {\scriptsize{}238.5} & {\scriptsize{}37.96} & {\scriptsize{}297.4} & {\scriptsize{}39.73} & {\scriptsize{}-}\tabularnewline
{\scriptsize{}Retina-I} & {\scriptsize{}36.3} & {\scriptsize{}119.2} & {\scriptsize{}30.31} & {\scriptsize{}246.1} & {\scriptsize{}28.49} & {\scriptsize{}1.39}\tabularnewline
{\scriptsize{}Retina-M} & {\scriptsize{}36.5} & {\scriptsize{}119.2} & {\scriptsize{}26.34} & {\scriptsize{}190.2} & {\scriptsize{}25.36} & {\scriptsize{}1.57}\tabularnewline
{\scriptsize{}Retina-F} & {\scriptsize{}36.8} & {\scriptsize{}119.2} & {\scriptsize{}26.18} & {\scriptsize{}254.1} & {\scriptsize{}30.08} & {\scriptsize{}1.32}\tabularnewline
{\scriptsize{}Retina-U} & {\scriptsize{}35.8} & {\scriptsize{}119.2} & {\scriptsize{}14.18} & {\scriptsize{}244.2} & {\scriptsize{}28.53} & {\scriptsize{}1.39}\tabularnewline
\hline 
{\scriptsize{}FSAF} & {\scriptsize{}37.4} & {\scriptsize{}205.5} & {\scriptsize{}36.41} & {\scriptsize{}283.1} & {\scriptsize{}37.10} & {\scriptsize{}-}\tabularnewline
{\scriptsize{}FSAF-I} & {\scriptsize{}37.2} & {\scriptsize{}102.7} & {\scriptsize{}28.52} & {\scriptsize{}231.5} & {\scriptsize{}26.89} & {\scriptsize{}1.38}\tabularnewline
{\scriptsize{}FSAF-M} & {\scriptsize{}37.3} & {\scriptsize{}102.7} & {\scriptsize{}24.09} & {\scriptsize{}171.3} & {\scriptsize{}23.00} & {\scriptsize{}1.61}\tabularnewline
{\scriptsize{}FSAF-F} & {\scriptsize{}37.6} & {\scriptsize{}102.7} & {\scriptsize{}23.19} & {\scriptsize{}233.6} & {\scriptsize{}27.90} & {\scriptsize{}1.33}\tabularnewline
{\scriptsize{}FSAF-U} & {\scriptsize{}36.4} & {\scriptsize{}102.7} & {\scriptsize{}19.23} & {\scriptsize{}242.2} & {\scriptsize{}27.99} & {\scriptsize{}1.33}\tabularnewline
\hline 
{\scriptsize{}ATSS$^{\star}$} & {\scriptsize{}38.1} & {\scriptsize{}204.4} & {\scriptsize{}32.29} & {\scriptsize{}283.1} & {\scriptsize{}37.14} & {\scriptsize{}-}\tabularnewline
{\scriptsize{}ATSS-I} & {\scriptsize{}37.9} & {\scriptsize{}102.2} & {\scriptsize{}24.48} & {\scriptsize{}234.6} & {\scriptsize{}26.76} & {\scriptsize{}1.39}\tabularnewline
{\scriptsize{}ATSS-M} & {\scriptsize{}38.0} & {\scriptsize{}102.2} & {\scriptsize{}22.20} & {\scriptsize{}177.1} & {\scriptsize{}23.72} & {\scriptsize{}1.57}\tabularnewline
{\scriptsize{}ATSS-F} & {\scriptsize{}38.3} & {\scriptsize{}102.2} & {\scriptsize{}21.54} & {\scriptsize{}236.9} & {\scriptsize{}28.18} & {\scriptsize{}1.32}\tabularnewline
{\scriptsize{}ATSS-U} & {\scriptsize{}36.7} & {\scriptsize{}102.1} & {\scriptsize{}12.41} & {\scriptsize{}228.5} & {\scriptsize{}26.35} & {\scriptsize{}1.41}\tabularnewline
\hline 
{\scriptsize{}PAA$^{\star}$} & {\scriptsize{}39.0} & {\scriptsize{}204.4} & {\scriptsize{}32.29} & {\scriptsize{}283.1} & {\scriptsize{}37.16} & {\scriptsize{}-}\tabularnewline
{\scriptsize{}PAA-I} & {\scriptsize{}39.4} & {\scriptsize{}102.2} & {\scriptsize{}24.82} & {\scriptsize{}234.7} & {\scriptsize{}27.29} & {\scriptsize{}1.36}\tabularnewline
{\scriptsize{}PAA-M} & {\scriptsize{}39.4} & {\scriptsize{}102.1} & {\scriptsize{}23.00} & {\scriptsize{}174.9} & {\scriptsize{}23.30} & {\scriptsize{}1.59}\tabularnewline
{\scriptsize{}PAA-F} & {\scriptsize{}39.6} & {\scriptsize{}102.1} & {\scriptsize{}21.95} & {\scriptsize{}235.5} & {\scriptsize{}27.81} & {\scriptsize{}1.34}\tabularnewline
{\scriptsize{}PAA-U} & {\scriptsize{}38.5} & {\scriptsize{}102.2} & {\scriptsize{}13.19} & {\scriptsize{}228.4} & {\scriptsize{}26.39} & {\scriptsize{}1.41}\tabularnewline
\hline 
{\scriptsize{}Faster} & {\scriptsize{}37.4} & {\scriptsize{}199.3} & {\scriptsize{}41.75} & {\scriptsize{}294.6} & {\scriptsize{}44.28} & {\scriptsize{}-}\tabularnewline
{\scriptsize{}Faster-M} & {\scriptsize{}37.8} & {\scriptsize{}99.55} & {\scriptsize{}30.96} & {\scriptsize{}197.0} & {\scriptsize{}25.58} & {\scriptsize{}1.73}\tabularnewline
{\scriptsize{}Faster-M} & {\scriptsize{}36.6} & {\scriptsize{}49.82} & {\scriptsize{}17.48} & {\scriptsize{}130.8} & {\scriptsize{}14.43} & {\scriptsize{}3.07}\tabularnewline
{\scriptsize{}Faster-F} & {\scriptsize{}37.8} & {\scriptsize{}99.70} & {\scriptsize{}25.95} & {\scriptsize{}240.6} & {\scriptsize{}28.79} & {\scriptsize{}1.54}\tabularnewline
{\scriptsize{}Faster-U} & {\scriptsize{}33.5} & {\scriptsize{}99.90} & {\scriptsize{}9.861} & {\scriptsize{}224.8} & {\scriptsize{}28.06} & {\scriptsize{}1.58}\tabularnewline
\end{tabular}{\scriptsize\par}
\end{spacing}
\end{table}

\subsection{Prune for Detection}

Pruning object detection is more challenging than image classification
due to its larger input size and more complicated networks, which
demands model pruning more than image classification. Besides, many
pruning methods based on BN scaling parameters can not be directly
applied. However, our method can not only be applied to image classification,
but also object detection, thanks to its general importance estimation
and the proposed layer grouping for pruning coupled channels. We
prune one-stage methods including RetinaNet \cite{lin2017focal},
FSAF \cite{zhu2019feature}, ATSS \cite{zhang2020bridging} and PAA
\cite{paa2020}, and two-stage method Faster R-CNN \cite{ren2015faster}
to extensively validate the effectiveness of our method. We present
the pruning results of our method for various detection frameworks
in Fig. \ref{fig:regnet} (d) and Tab. \ref{tab:det}. Similar to
pruning classification models, normalizing importances with memory
reduction and pruning coupled channels together (``-M'') leads to
the highest efficiency. Our method effectively prunes detection networks
without losing average precision, in some cases our pruned model even
receives higher mAP than the unpruned baseline. More importantly,
our method delivers practical inference speedup, \eg, we achieve
a $3\times$ speedup by pruning Faster R-CNN with only 0.8\% mAP drop.
We also compare the pruned networks with state-of-the-art method Slimmable
Networks \cite{yu2018slimmable} in Tab. \ref{tab:slim}, as shown
we can achieve higher mAP, lower mAP drop under comparable or less
FLOPs.
\begin{table}
\caption{\label{tab:slim}{\small{}Compare with Slimmable Networks for pruning
Faster-RCNN on COCO. AP for pruned model and the unpruned model (B-AP),
and the AP drop ($\triangle$) are shown. ``F'' denotes the percentage
of remained FLOPs.}}

\begin{spacing}{1.0}
\centering{}{\small{}}%
\begin{tabular}{c|cccc}
{\small{}model} & {\small{}AP(\%)} & {\small{}B-AP(\%)} & {\small{}$\triangle$(\%)} & {\small{}F(\%)}\tabularnewline
\hline 
{\small{}Slim \cite{yu2018slimmable}} & {\small{}36.1} & {\small{}36.4} & {\small{}0.3 $\downarrow$} & {\small{}56}\tabularnewline
{\small{}Ours} & {\small{}37.8} & {\small{}37.4} & {\small{}0.4 $\uparrow$} & {\small{}50}\tabularnewline
\hline 
{\small{}Slim \cite{yu2018slimmable}} & {\small{}34.0} & {\small{}36.4} & {\small{}2.4 $\downarrow$} & {\small{}25}\tabularnewline
{\small{}Ours} & {\small{}36.6} & {\small{}37.4} & {\small{}0.8 $\downarrow$} & {\small{}25}\tabularnewline
\end{tabular}{\scriptsize\par}
\end{spacing}
\end{table}

\begin{figure}
\begin{centering}
\includegraphics[scale=0.21]{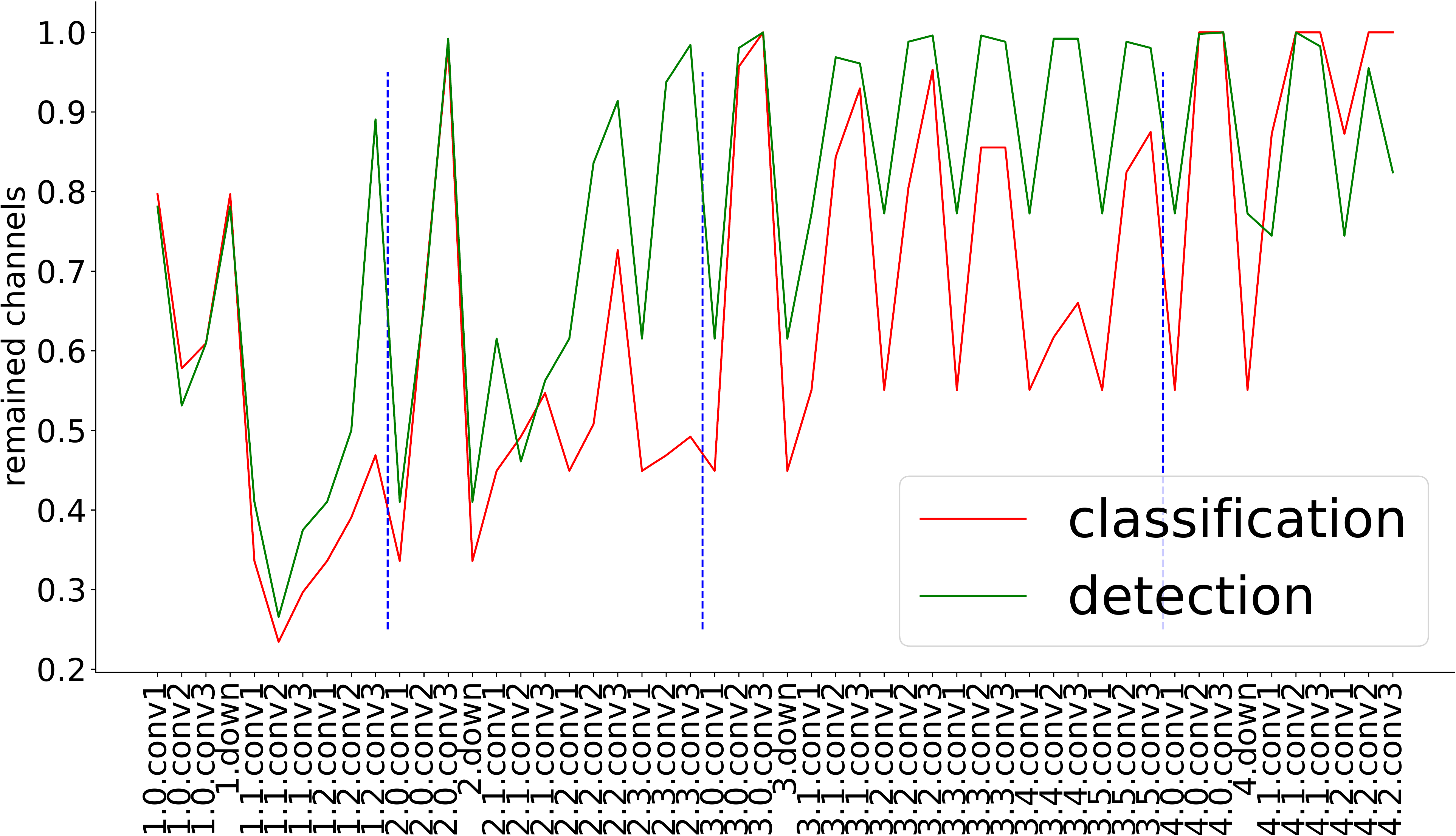}
\par\end{centering}
\caption{\label{fig:det}{\small{}The percentage of remained channels in the
pruned backbone over different layers for classification and detection.
Features maps are $2\times$ down-sampled at blue dashed lines.}
}
\end{figure}

Considering the intrinsic differences between image classification
and object detection, we compare the pruned network structures between
them. As in Fig. \ref{fig:det}, we find that the pruned classification
network keeps more capacity in later stages where the spatial resolution
is rather small, as classification needs more global features. However,
for detection the early stages also remain a large portion of channels,
as detection should extract features at different scales to detect
objects with various sizes. This validates that our method can be
adaptively applied to different tasks.

\section{Conclusion}

We present a general channel pruning framework for complicated structures.
We propose the layer grouping algorithm to find coupled channels and
make them share the binary mask. Based on the single-channel importance
approximated by Fisher information, we compute the overall importance
of coupled channels by the chain rule of gradient computation. We
prune the coupled channels simultaneously for better accuracy-efficiency
trade-off. Moreover, normalizing channel importances by memory reduction
rather than FLOPs is proposed to deliver more speedup. Extensive experiments
on pruning various network structures with residual connections, GConv/DWConv
and FPN in detection are explored and verify the effectiveness.

Inspired by
the memory-bound nature of GPUs, we propose to normalize channel importance
by memory reduction, which can bring a better trade-off between accuracy
and speedup. In future work we will theoretically model the relationships
between core factors (\emph{e.g.,} FLOPs, memory) and inference speed
on various platforms (\emph{e.g.,} GPU/CPU/TPU).

\section*{Acknowledgements}
The work described in this paper was partially supported by the Natural Science Foundation of Guangdong Province (No. 2020A1515010711), the Special Foundation for the Development of Strategic Emerging Industries of Shenzhen (No. JCYJ20200109143010272 and No. JCYJ20200109143035495), the Innovation and Technology Commission of the Hong Kong Special Administrative Region, China (Enterprise Support Scheme under the Innovation and Technology Fund B/E030/18) and the Shanghai Committee of Science and Technology, China (Grant No. 20DZ1100800).


\bibliography{example_paper}
 \bibliographystyle{icml2021}


\appendix

\end{document}